\documentclass[conference]{IEEEtran}
\IEEEoverridecommandlockouts
\usepackage{cite}
\usepackage{amsmath,amssymb,amsfonts}
\usepackage{algorithmic}
\usepackage{graphicx}
\usepackage{textcomp}
\usepackage{xcolor}
\usepackage{url}
\usepackage{multirow}
\usepackage{subfig} 
\def\BibTeX{{\rm B\kern-.05em{\sc i\kern-.025em b}\kern-.08em
    T\kern-.1667em\lower.7ex\hbox{E}\kern-.125emX}}
\begin{document}

\title{A Computer Vision Approach for Autonomous Cars to Drive Safe at Construction Zone}

\author{\IEEEauthorblockN{Abu Shad Ahammed}
\IEEEauthorblockA{\textit{Chair of Embedded Systems} \\
\textit{University of Siegen}\\
Siegen, Germany \\
ORCID: 0009-0007-6715-8098}
\and
\IEEEauthorblockN{Md Shahi Amran Hossain}
\IEEEauthorblockA{\textit{Chair of Embedded Systems} \\
\textit{University of Siegen}\\
Siegen, Germany \\
ORCID: 0009-0000-9466-0912}
\and
\IEEEauthorblockN{Roman Obermaisser}
\IEEEauthorblockA{\textit{Chair of Embedded Systems} \\
\textit{University of Siegen}\\
Siegen, Germany \\
ORCID: 0009-0002-4483-1503}

}

\maketitle

\begin{abstract}
To build a smarter and safer city, a secure, efficient, and sustainable transportation system is a key requirement. The autonomous driving system (ADS) plays an important role in the development of smart transportation and is considered one of the major challenges facing the automotive sector in recent decades. A car equipped with an autonomous driving system (ADS) comes with various cutting-edge functionalities such as adaptive cruise control, collision alerts, automated parking, and more. A primary area of research within ADAS involves identifying road obstacles in construction zones regardless of the driving environment. This paper presents an innovative and highly accurate road obstacle detection model utilizing computer vision technology that can be activated in construction zones and functions under diverse drift conditions, ultimately contributing to build a safer road transportation system. The model developed with the YOLO framework achieved a mean average precision exceeding 94\% and demonstrated an inference time of 1.6 milliseconds on the validation dataset, underscoring the robustness of the methodology applied to mitigate hazards and risks for autonomous vehicles.
\end{abstract}


\section{Introduction}
\label{sec:int}
An automated or autonomous driving system refers to a complex combination of hardware and software components collectively responsible for controlling a vehicle's operation without human intervention. In this type of system, electronic and mechanical components take over the tasks of perception, navigation, and other necessary driving gestures, e.g. acceleration, braking, and steering of the vehicle instead of a human driver. ADS seeks to automate the longitudinal (acceleration and braking) and lateral (steering) movements of vehicles, minimizing the need for human involvement by using sensors, cameras, radars and artificial intelligence \cite{de2014effects}. A smart and secure automated driving system integrated with a car has the potential to improve traffic efficiency, reduce traffic jams, and increase road safety. The object detection system within automated driving plays a crucial role in enabling vehicles to drive safely and avoid potential hazards. This type of system is created using computer vision (CV) algorithms, which allow vehicles to 'perceive' and understand their environment by processing visual information from cameras and various sensors to identify obstacles such as other cars, pedestrians, animals, and different objects on the road. The research presented in this paper is focused on the detection of road obstructions such as beacons, barricades, and cones that are found specially in construction zones using CV.\\
Artificial intelligence (AI) has reformed the automotive industry and has become a decisive technology for the development of autonomous vehicles and the advanced driver assistance system (ADAS). Machine learning and computer vision algorithms are part of AI and are often utilized in the automotive industry to construct a decision-making system within automated driving to determine appropriate driving behaviors based on environmental information and vehicle status \cite{he2023robust}. This system acts as the "brain" of the automated driving system, ensuring safe and efficient driving operations e.g. object detection and recognition, lane detection and keeping, traffic sign recognition, automatic emergency braking, parking assistance, road environment monitoring, etc. \cite{hossain2024impact}. Computer vision algorithms, a significant part of decision-making systems, use data from integrated automotive sensors, radars, and lidars to evaluate specific detectable parameters, subsequently applying these insights to prediction or decision-making processes. In recent times, leading automobile manufacturers have recognized CV as essential for autonomous driving, particularly for addressing issues such as the sudden emergence of unforeseen obstacles that require alterations in the driving route and direction from the natural environment. Such scenarios are common in the construction zone, which is an area where road work takes place and can involve lane closures or detours due to the presence of heavy machinery, construction materials and even road workers in close proximity \cite{awolusi2019active}. Modern cars equipped with the latest ADS have the ability to drive safely in the construction zone acknowledging the changing environment. But because of the presence of data drift caused by car hardware or road environment, there exists a serious threat for ADS by degrading the CV-based object detection and recognition model. Several factors can cause a CV model to under-perform, specially when the data set is insufficient, imbalanced, noisy, or lacks diversity, it can lead to significant data drift in practical simulations, thereby deteriorating the model's performance. The study outlined in this article underscores the limitations of the vision model, in particular its reduced effectiveness in identifying objects such as construction zone obstacles under deteriorated or drifted traffic conditions, and proposes a solution by developing a custom CV module with an annotated and drifted dataset generated from the CARLA simulator.\\ 
The obstruction detection model in the construction zone employs YOLO version 8, which is well known for its precision and rapid inference time. Three obstacles have been considered that are often found in construction zones- beacon, cone, and barrier. For preparing the drifted data set, images are generated from the CARLA simulator while considering several factors such as different city maps, diverse road layouts, different lighting and weather conditions to create a robust and comprehensive data set. In addition to detailing the model development procedures, the paper also contains a concise performance evaluation of the developed module based on selected evaluation criteria.\\
The remainder of the manuscript is organized as follows. Section \ref{lit} provides a meta-analysis of existing research works, methods, and techniques by different authors. Section \ref{odd} defines the operational design and domain (ODD) of the construction zone where CV model should function. Section \ref{hara} explains the possible hazards and risk in construction zone. A conceptual overview of data generation with CARLA simulator and preprocessing steps are presented in Section \ref{drift}. The development of the CV module with YOLO and selected hyperparameters is discussed in Section \ref{obj}. Section \ref{res} presents a thorough performance evaluation of the object detection model, including performance and loss graphs. Finally, Section \ref{conc} provides a conclusion, mentioning the existing challenges and a look at the possibilities in the future.
\section{Related Research}
\label{lit}
A unique aspect of the research demonstrated in this paper is its policy on utilizing simulated data, given the absence of real data storage for obstacles in various road construction scenarios and drift conditions.  Computer vision algorithms are quite famous for their ability to detect objects, which is an important aspect of this research to investigate how they work on simulated construction zone data in a drifted scenario. Although earlier research has explored the impact of such data drifts on vision models, this study specifically targets its effects on the riskier construction zone, even in deteriorated road environments to ensure safe driving.\\
AI algorithm is considered as a key aspect of ensuring the safety and efficiency of autonomous vehicles and advanced driver assistance systems. In many past research works \cite{bertozzi1998gold, levi2015stixelnet}, the use of AI algorithms such as deep convolution networks or vision-based architectures was highlighted for object detection and road segmentation, crucial for applications of autonomous driving and vehicle safety. As mentioned in \cite{alam2022object}, AI can support numerous countries around the world that experience a significant impact on their transportation systems due to inadequate advanced technology, inadequate infrastructure, violation of traffic regulations, and unsafe driving practices, among other factors. Qi et al. \cite{qi2019convolutional} focus on a convolutional neural network approach for detecting and assessing environmental obstacles during vehicle operation, emphasizing the need for highly accurate and instantaneous methods to prevent traffic collisions and ensure safety. Together, these sources underscore the importance of cutting-edge computer vision techniques in managing road hazards and improving driver security. However, the performance of such algorithms often gets deteriorated due to drifted scenarios. The author of this paper had already done a research \cite{hossain2024impact} that provided a thorough analysis on how data drift may affect the performance of the traffic sign detection model developed with the computer vision algorithm. The negative impact of data drift was mitigated by training the original detection model with a data set created in drifted scenarios, leading to enhanced safety critical automotive systems. After training the CV model with the drifted data set, it showed a higher precision of 97. 5\% compared to a precision of 58. 27\% of the model trained with normal road environment data.\\
In our study, we used YOLO, which is a popular and efficient object detection model in computer vision and is often used in the automotive sector for its fast inference capability. In our exploration a significant amount of research work \cite{sivaraman2013looking, sun2022rsod, lan2018pedestrian} has been found that utilized the YOLO model for various purposes like monitoring road traffic, understanding vehicle navigation behavior, pedestrian detection, etc. Meng et al. in their paper from 2023 \cite{meng2023yolov5s} proposed a foggy weather detection method based on the YOLOv5s framework, named YOLOv5s-Fog. Foggy weather is a classic example of drift situation considered in their model by integrating a novel target detection layer called SwinFocus. Compared to the baseline model, YOLOv5s, their YOLOv5s-Fog model achieves a 5. 4\% increase in mean average precision in the RTTS dataset, reaching 73.4\%.
\section{Construction Zone ODD}
\label{odd}
\begin{figure}[!b]
    \centering
    \subfloat[Barrier]{\label{fig:std1}
        \includegraphics[width=0.24\textwidth]{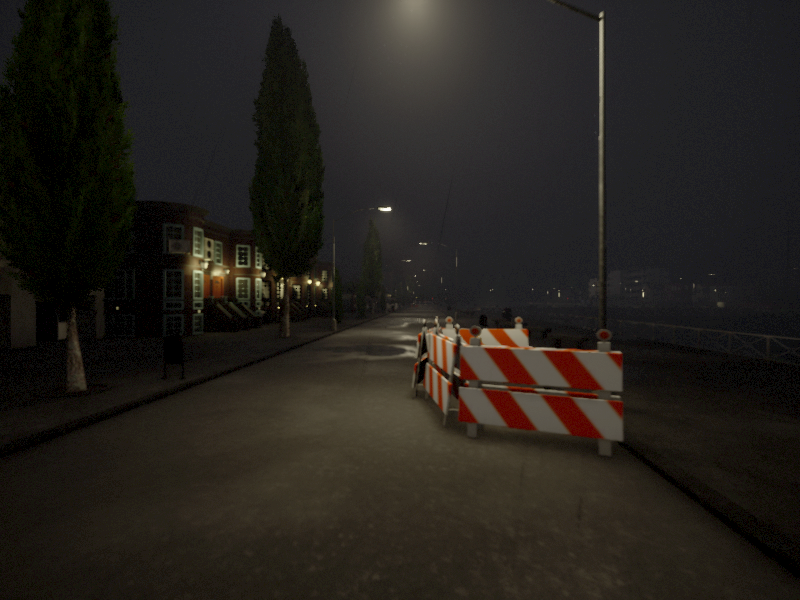}}
    \hspace{1cm} 
    \subfloat[Beacon]{\label{fig:drift2}
        \includegraphics[width=0.24\textwidth]{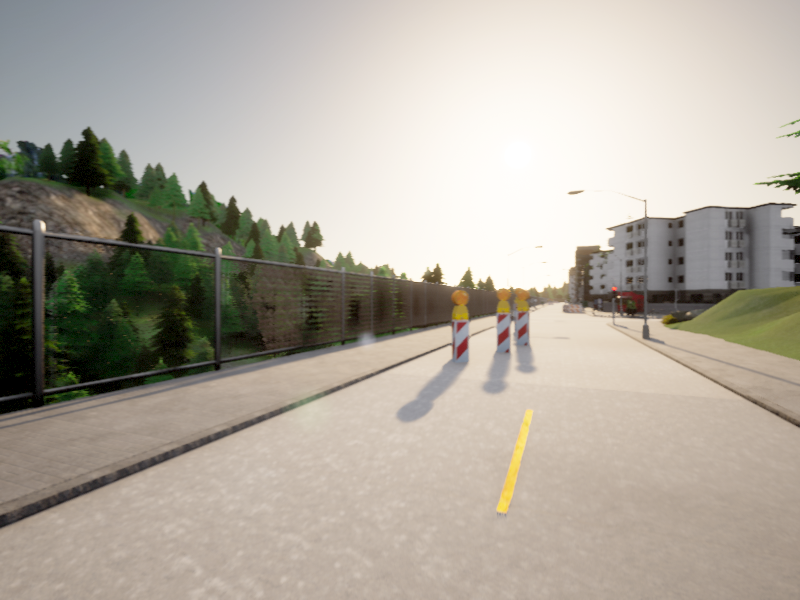}}
    \hspace{1cm}
    \subfloat[Cone]{\label{fig:drift3}
        \includegraphics[width=0.24\textwidth]{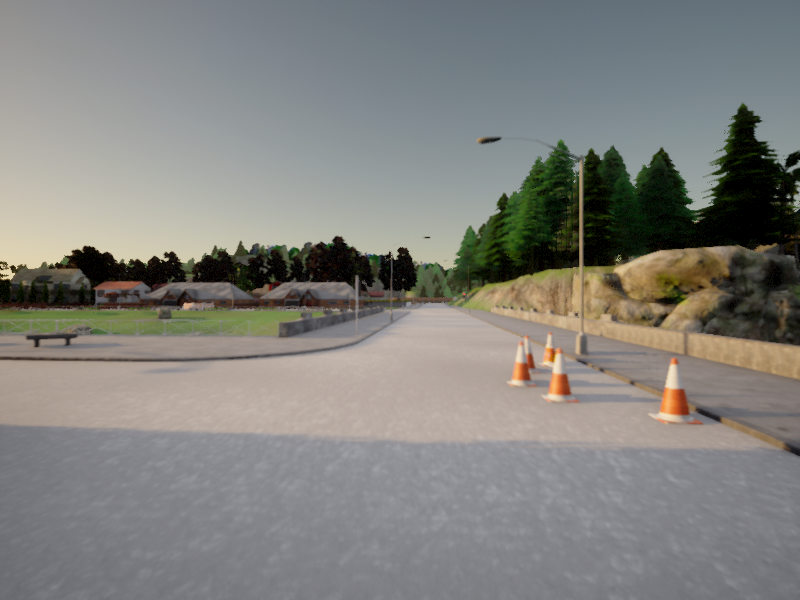}}
    \caption{Road Obstructions in Construction Zones}
    \label{fig:drift}
\end{figure}

The operational design and domain (ODD) refers to all the conceivable conditions, scenarios, and restrictions under which an automated driving system or autonomous vehicle is designed to operate safely \cite{garcia2022vehicle}. It specifies the limits within which the system can operate efficiently and encompasses elements such as environmental conditions, geographical areas, time of day, types of roads, speed regulations, and other functional constraints. When incorporating AI algorithms such as CV for an autonomous driving system, it is urgent to define the operating conditions under which the system will operate and ensure only within this perimeter the safety of the system. Otherwise, the intrinsic uncertainty associated with AI-generated results and the erratic actions of other agents within the operational domain amplify the complexity of these systems \cite{adedjouma2024defining}.\\
Driving in construction zones can pose multiple challenges that can increase the risk of accidents for both motorists and construction workers. Some of the primary challenges are narrowed lanes and traffic pattern changes, limited visibility,  distracted driving, increased risk of collisions, construction hazards lying on the road, etc. In addition, behaviors such as speeding, tailgating, and abrupt lane changes are significant causes of accidents in construction zones. Before developing such a CV-based model that can effectively mitigate these challenges for safe navigation, an ODD for the construction zone must be defined to include specific criteria that the model should acknowledge. In our study, the following criteria were listed to generate data sets representing the construction zone:
\begin{itemize}
    \item Road configuration including detours, altered or temporary lane markings, and narrowed lanes
    \item Variability in traffic patterns that includes different speed limit, vehicle navigation sign, stop and go signs
    \item Three types of road obstacles generally found in construction zone areas have been defined- cones, beacons, road barriers as can be seen in Fig \ref{fig:drift}
    \item Dynamic road layouts including city, village or highway
    \item Most importantly, drifted situations have been introduced such as poor day light, rainy/foggy weather, blurry visibility due to hardware failure, etc
\end{itemize}
\section{HARA in Construction Zone}
\label{hara}
Hazard Analysis and Risk Assessment (HARA) is a systematic process used in the automotive sector, particularly in the context of ISO 26262 concerning functional safety for road vehicles, to detect potential hazards, assessing associated risks, and implementing measures to mitigate these risks. HARA plays a vital role in ensuring that safety-critical vehicle systems, like those used in autonomous driving and Advanced Driver Assistance Systems (ADAS), are designed and implemented to minimize risks to an acceptable threshold \cite{khastgir2017towards}. Based on our research exploration, we identified the following hazards pertained to construction zones:
\begin{itemize}
    \item Risk of collisions with the road barrier intended to convey lane closures, shifts, and merging lanes due to not braking the car in right time
    \item High possibility of accidents with construction worker or their equipment if the beacons used for demarcation and protection of road workers are not detected
    \item Inadequate or unclear signage can deteriorate the ADAS functionality in autonomous vehicles leading to delayed deceleration
    \item Drift situation which is the prime focus in our research causes reduced visibility of road obstructions due to the operational ineffectiveness of sensors or cameras integrated with cars
\end{itemize}
The traditional ADAS system which combine data from multiple sensors to detect obstacles may fail to do so in drift situations caused by dust, rain or low light. This encourages the development of our specialized computer vision model that takes into account sensor limitations and adverse weather conditions while performing semantic segmentation on the front view to categorize each pixel of an image and identify trained objects.
\section{Data Organization and Pre-processing}
\label{drift}
Data organization and pre-processing in the fundamental step for computer vision tasks such as image detection and object recognition. Effective data preprocessing methods are significant for boosting input data quality, enhancing model accuracy, and expedite the training phase \cite{tribuana2024image}. The data sets planned in our research are standard and drift images of traffic scenarios directly generated from CARLA. CARLA, the abbreviated form of Car Learning to Act, is an unreal engine-based open-source autonomous driving simulator consisting primarily of two modules, the simulator and the Python/C++ API module. It was born from a joint initiative of researchers from Intel Labs, Toyota Research Institute, and the Computer Vision Center of Barcelona and was released in 2017 \cite{ly2020learning}. CARLA has a high-fidelity 3D simulation environment that includes urban, suburban, and highway settings. The environment also offers dynamic illumination, and variable weather conditions with a wide array of scenarios for evaluating autonomous driving systems. As mentioned in \cite{pmlr-v78-dosovitskiy17a}, the modular architecture of CARLA consist of a server and client architecture where the server handles the simulation's core functions, including sensor rendering, physics computation, and updates to the world state. The client, which can be controlled via APIs, allows developers to interact with the simulator, control vehicles, access sensor data, and create custom scenarios for testing and training autonomous driving algorithms. One of CARLA's significant benefits is its array of automotive sensors like cameras, lidar, radar, IMU, etc., which are typically integrated into the construction of autonomous vehicles. From the client side, these sensors can be placed anywhere on the car, allowing users to simulate various configurations and evaluate sensor fusion techniques.\\
Firstly, to create the image data sets, the mentioned road obstructions were positioned on eight predefined simulator maps. All the images were generated from the camera mounted on the front end of the vehicle to imitate the driver's point of view. These image data can also be collected with a camera from the actual road environment. But this process requires a lot of time and labor, as well as a lot of risk factors compared to the way we used the CARLA simulator, as it is an open source environment, offering more flexibility and scalability to support the development of digital assets. Initially, we created a data set consisting of 4,500 images including drift conditions. In the following subsections, the data management and preprocessing steps followed are briefly explained. 

\subsection{Data Acquisition from CARLA}
To acquire the data from the CARLA simulator, the first step was to set up the simulation environment with custom settings and traffic after launching and connecting to the CARLA client server. Next, we chose maps from the simulator's built-in list to obtain different environments such as urban streets, highways, and rural roads. Following customization were done to define the driving environment:
\begin{itemize}
    \item Set up an ego vehicle with Python API that will roam around the selected maps. The vehicle was configured to manual so that client side could have better control
    \item Customized the weather conditions such as sun position, fog density, wetness, etc. to create diverse datasets that reflect real-world conditions. To simulate the day-light condition and sun light reflection to the perceived obstacles, the sun position was customized through the Python API
    \item The ego vehicle was outfitted with multiple sensors, including cameras, lidar, and radar. However, the primary focus was to gather image data using the front-mounted RGB camera on the vehicle
    \item The three obstacles were custom-created, modified, and placed on all maps manually through Unreal Engine 4. Because these static actors cannot be generated or controlled by the Python API
    \item A robot operating system 2 (ROS2) bridge was used to maintain standard communication between CARLA and ROS2 for the purpose of collecting simulation data. A custom-configured publisher node on that bridge was responsible for continuously publishing image data from the RGB camera while the ego vehicle was moving. Subsequently, a synchronous subscriber node was implemented to extract and export the image. The entire procedure is managed and executed using Docker to prevent dependency issues
\end{itemize}

\subsection{Data Filtering}
After completing the acquisition process, all recorded images were manually checked to filter the images that had selected obstacles, while all the others without obstacles were discarded. During this process, some of the filtered images in which the obstacles were very close to the camera were removed. Because the goal is to detect obstacles in a range where autonomous vehicles will have time to navigate or decelerate after detecting them.

\subsection{Data Augmentation and Labeling}
The filtered data set was not initially balanced and lacked diversity, which is a weakness to create a robust CV model. As a solution, we have employed techniques for data augmentation, such as mirroring, blurring, flipping, skewing, and rotating. To include drift scenarios, the Python API was used to simulate drifted weather conditions as discussed in Section \ref{hara}. The drift was adjusted following below procedures:
\begin{itemize}
    \item Color space transformation by adjusting brightness, contrast, saturation and hue
    \item Injected Gaussian noise and blur effect to imitate ineffectiveness of camera sensor
    \item The obstacles were modified by their size and position at the road
\end{itemize}
Afterwards, the developed image date set was labeled using the CVAT (Computer Vision Annotation Tool) with their corresponding categories. CVAT, developed by Intel, is a free, open-source, web-based image and video annotation tool used to label data for computer vision algorithms \cite{musleh2023image}. We uploaded the drift data set to CVAT after creating a task. There are several annotation tools for labeling objects in the images where we chose boundary boxes in rectangular shape. After each annotation, we selected the respective class before saving the task. Once the annotations were completed, the dataset was exported in YOLO format containing two folders. One was the folder with images and another with '.txt' files that contained details about the annotated objects, such as their coordinates, class labels, and other attributes.

\subsection{Data Segmentation}
Data segmentation is a necessary step to avoid over-fitting or under-fitting and to accurately assess the performance of the computer vision model. To do that, the final image data set was broken into three distinct splits: train, test, and validation, where significant effort was put into having adequate variations for all environmental aspects and keeping the data balanced. Table \ref{tab:split} shows the length of the data set with each obstacle category.
\begin{table}[!b]
\renewcommand{\arraystretch}{1.3}
\caption{\textsc{Length of Data Split}}
\centering
\label{tab:split}
\begin{tabular}{|l|l|l|l|}
\hline
\multicolumn{1}{|c|}{\textbf{Type}} & \multicolumn{1}{c|}{\textbf{Beacon}} & \multicolumn{1}{c|}{\textbf{Cone}} & \multicolumn{1}{c|}{\textbf{Barrier}} \\ \hline
Train                               & 1126                                 & 530                                & 1072                                  \\ \hline
Validation                          & 240                                  & 120                                & 240                                   \\ \hline
Test                                & 10                                   & 10                                 & 10                                    \\ \hline
\end{tabular}
\end{table}

\begin{figure}[!t]
    \center
    \includegraphics[width=0.5\textwidth]{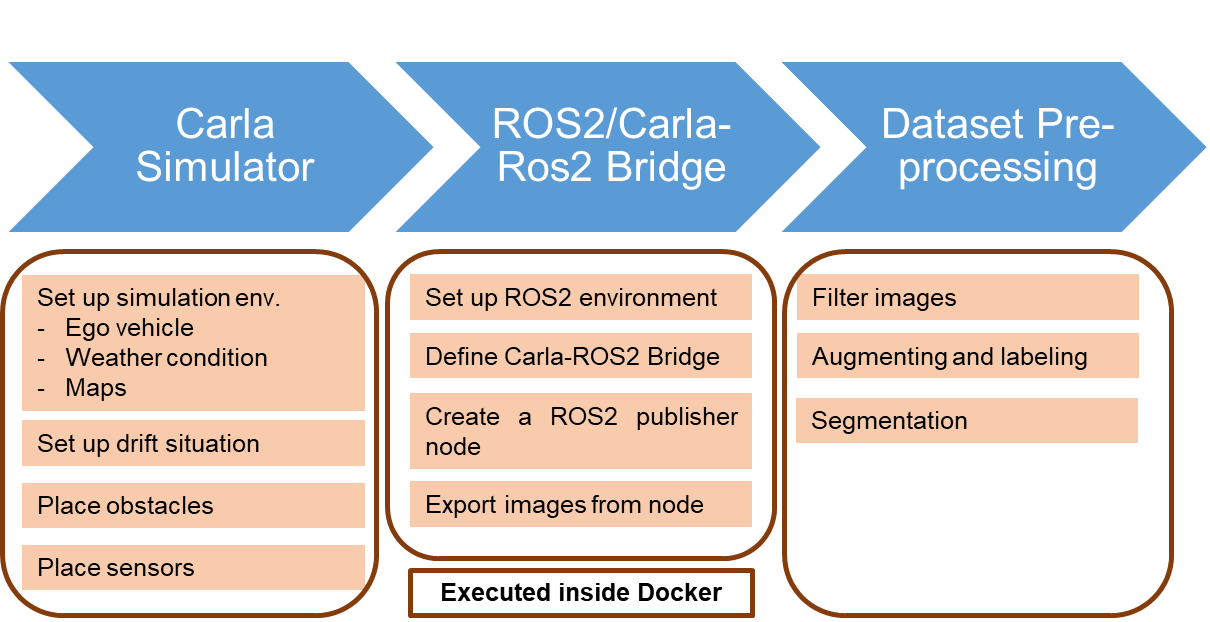}
    \caption{An Overview of Data Management and Pre-processing}
    \label{fig:prep} 
    \end{figure}

The summary of the data management steps mentioned above is shown in Fig. \ref{fig:prep}.
\section{Object Detection Modeling for CZ}
\label{obj}
There were several methods in consideration to develop the computer vision model e.g., YOLO, R-CNN, SSD, and Faster R-CNN. Each of these methods offer distinct preferences and drawbacks, which makes them suitable for diverse applications. In any case, the YOLO (You Only Look Once) framework has been a major player in this field, particularly with its most recent versions because of its combination of real-time detection and high precision. For this research, YOLOv8 was chosen instead of its successor YOLOv9 because YOLOv8 has the better ability to segment objects at the pixel level \cite{sapkota2024comparing}, which is invaluable for tasks like semantic segmentation and object detection in drift situation. In addition, YOLOv9 focuses more on deeper networks, additional parameters, and fine-grained feature extraction, which increase model complexity, overfitting, and diminishing returns, leading to higher computational costs without significant gains in accuracy \cite{folio3YOLOv9YOLOv8, encordComparativeAnalysis}. And unlike its predecessors, YOLOv8 underwent a "full-scale reloading", with significant changes to its network architecture and training strategy, resulting in improved capabilities on a variety of object detection tasks. In addition, it has a more robust loss function and sophisticated data enhancement techniques, which contribute to its improved detection accuracy and generalization capabilities.\\
The first step in creating the YOLO-based road obstacle detection model involved commencing the training phase using the annotated data set downloaded from CVAT. Ultralytics, which provides a simple and efficient way to train YOLOv8, was used to complete the training process, and before starting the actual training, a '.yaml' file was created and configured according to the Ultralytics documentation \cite{UltralyticsTrain2024}. The '.yaml' file included the path to training, validation, and testing images. In addition, the three classes corresponding to the obstacles were described there. Class '0' represented cones, class '1' road barriers, and class '2' beacons. Python 3.11 was used to configure the environment required to execute the scripts. At first an environment is set up using required libraries such as torchaudio, opencv, ultralytics, etc. The ultralytics Python library provided access to use the custom pre-trained YOLOv8 model which we used to train our custom model. The hyperparameters applied during the training phase are as follows:
\begin{itemize}
    \item Optimizer: Adamw
    \item Learning rate: 0.000833
    \item Epochs: 100
    \item Confidence score: 0.2
    \item Image size: 640 
\end{itemize}
The training steps were followed by model validation and testing to find out how the model performs and whether the inference time is suitable to use in practical scenarios. Details about the performance of our model are discussed in the next section.
\section{Results and Discussion}
\label{res}
\begin{figure}[!t]
\center
\includegraphics[width=0.50\textwidth]{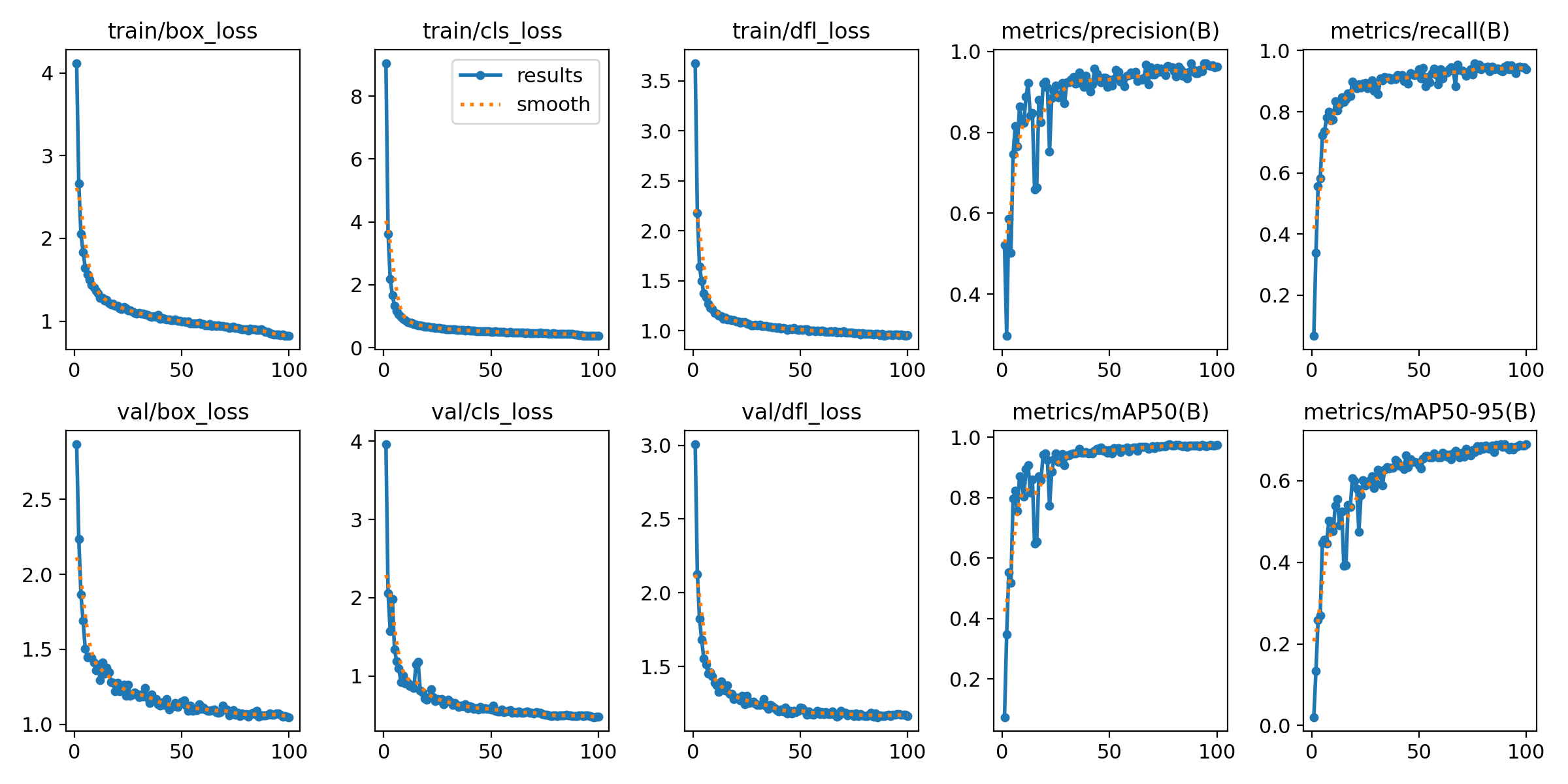}
\caption{Performance Curves of the Developed Model}
\label{fig:OrgRes} 
\end{figure}

Some basic metrics were used to assess the performance of the YOLO model in recognizing and classifying traffic signs as below \cite{quach2023evaluating}:
\begin{itemize}
\item Precision: Precision quantifies the ratio of accurately predicted positive instances to the total number of positive predictions made by the model
\item Recall: Recall assesses the ratio of correctly identified positive instances to the total number of actual positive cases in the dataset
\item F1 Score: It is a weighted harmonic mean of precision and recall which gives insight into both the correctness of the positive prediction and the ability to find all the positive instances
\item Mean Average Precision 50 (mAP50): Assesses the model's ability to detect objects effectively. It determines the average of precision-recall curves for all classes at a 50\% Intersection over Union (IoU) criterion
\item Mean Average Precision 50-95 (mAP50-95): Offers a more comprehensive and stringent evaluation of the performance of the model. The IoU threshold varies from 50\% to 95\%, and the average precision-recall curves are computed for all categories
\end{itemize}
After training, the YOLO model generally produces several performance curves demonstrating confusion matrices, precision-recall, F1 score, and losses. The curves offer insight into multiple dimensions of the model's accuracy, efficiency, and ability to generalize. In Figures \ref{fig:OrgRes}, autogenerated curves are shown to apprehend the performance of the model.
The top row in figure indicates epoch-wise performance of the model during training. The bottom row represents the validation performance. Values demonstrated there were the average of all classes. The three loss curves for both the training and validation period showed a common behavior of starting high then decreases rapidly over the first few epochs, stabilizing in a much lower value. This indicates the model is always getting better at predicting and classifying the objects confined in bounding boxes. The precision and recall values at the validation instance although started low but then stabilizes to a higher values of 0.94 and 0.95 respectively. The mean average precision value during validation increased quickly and stabilizing near 0.97, indicating strong detection performance. mAP50-95 curve showed improvement over time, stabilizing around 0.69, suggesting the model maintains good performance across different IoU thresholds. A summarization of the object-wise model's performance is presented in Table \ref{tab:vperf} and \ref{tab:tperf}.
\begin{table}[!b]
\renewcommand{\arraystretch}{1.3}
\caption{\textsc{Performance Over Validation Data}}
\centering
\label{tab:vperf}
\begin{tabular}{|l|c|c|c|c|c|}
\hline
\multicolumn{1}{|c|}{\textbf{Class}} & \multicolumn{1}{c|}{\textbf{Precision}} & \multicolumn{1}{c|}{\textbf{Recall}} & \multicolumn{1}{c|}{\textbf{mAP50}} & \multicolumn{1}{c|}{\textbf{mAP50-95}} & \textbf{\begin{tabular}[c]{@{}c@{}}Inference\\ (ms)\end{tabular}} \\ \hline
Cone                                 & 0.97                                    & 0.92                                 & 0.97                                & 0.66                                   & \multirow{3}{*}{1.6}                                              \\ \cline{1-5}
Barrier                              & 0.93                                    & 0.96                                 & 0.98                                & 0.73                                   &                                                                   \\ \cline{1-5}
Beacon                               & 0.94                                    & 0.96                                 & 0.97                                & 0.75                                   &                                                                   \\ \hline
\end{tabular}
\end{table}
\begin{table}[!b]
\renewcommand{\arraystretch}{1.3}
\caption{\textsc{Performance Over Test Data}}
\centering
\label{tab:tperf}
\begin{tabular}{|l|c|c|c|c|c|}
\hline
\multicolumn{1}{|c|}{\textbf{Class}} & \multicolumn{1}{c|}{\textbf{Precision}} & \multicolumn{1}{c|}{\textbf{Recall}} & \multicolumn{1}{c|}{\textbf{mAP50}} & \multicolumn{1}{c|}{\textbf{mAP50-95}} & \textbf{\begin{tabular}[c]{@{}c@{}}Inference\\ (ms)\end{tabular}} \\ \hline
Cone                                 & 1.00                                    & 1.00                                 & 1.00                                & 0.78                                   & \multirow{3}{*}{7.62}                                              \\ \cline{1-5}
Barrier                              & 1.00                                    & 1.00                                 & 1.00                                & 0.79                                   &                                                                   \\ \cline{1-5}
Beacon                               & 1.00                                    & 0.91                                 & 0.96                                & 0.75                                   &                                                                   \\ \hline
\end{tabular}
\end{table}
The class-wise performance stat shown in the table is aligned with the numbers depicted in the above mentioned curves. The inference time found is less than 2 milliseconds during validation which infers the model's suitability if used in real world.
\section{Conclusion}
\label{conc}
This research aims to deliver a safe driving solution for autonomous cars in construction zones. The object detection system achieved an average precision that exceeded 90\% even in drift scenarios, ensuring its overall safety and reliability. In the future, this model could be explored for deployment in programmable hardware, facilitating integration into automotive systems, and becoming an important component of safety-critical automotive architectures.

\section*{Acknowledgment}
This work has been funded by the Federal Ministry of Education and Research (BMBF) as part of AutoDevSafeOps (01IS22087Q).
\bibliographystyle{ieeetr}
\bibliography{reference}
\end{document}